%% file: main.tex
\documentclass[final]{nesy2026} 

\usepackage[T1]{fontenc}
\usepackage[utf8]{inputenc}
\usepackage{microtype}
\usepackage{mathtools}
\usepackage{booktabs}
\usepackage{multirow}
\usepackage{multicol}
\usepackage{tabulary}         
\usepackage{array}
\usepackage[table]{xcolor}
\usepackage{xspace}
\usepackage{enumitem}
\usepackage{subcaption}
\usepackage{wrapfig}
\hypersetup{colorlinks=true,
            linkcolor=black,
            citecolor=blue!60!black,
            urlcolor=blue!60!black,
            hypertexnames=false}
\usepackage[hyperpageref]{backref}
\renewcommand*{\backrefalt}[4]{%
  \ifcase #1\relax
  \or (cited on p.~#2)%
  \else (cited on pp.~#2)%
  \fi}
\usepackage[capitalise,nameinlink]{cleveref}

\input{macros}

\title[SoftReason]{SoftReason: A Fully Differentiable Neuro-\emph{Soft}-Symbolic
       Deductive Reasoning Architecture over High-Dimensional Perceptual Data}

\clearauthor{\Name{Wael AbdAlmageed} \Email{wabdalm@clemson.edu}\\
\addr{Holcombe Department of Electrical and Computer Engineering, Clemson University}}

\begin{document}
\maketitle

\begin{abstract}
In many reasoning problems, the premises are not observed as discrete symbols, but must be inferred from high-dimensional inputs.  Further, the predicate vocabulary, argument structure, and trusted  evidence are supplied by a Knowledge Graph (KG), or rule definitions.  Classical neuro-symbolic pipelines have a discrete interface between perception and deduction.  We present a neuro-\emph{soft}-symbolic architecture for differentiable deductive reasoning over latent perceptual facts and knowledge-provided predicates.   \ours{} removes the gradient gap by representing the deductive state as a local soft interpretation tensor over candidate constants and predicates.  Perception proposes probabilistic base facts, KG triples enter as high-confidence soft evidence, and every query anchor, predicate choice, and closure update remains differentiable.  Our core innovation  is a learned differentiable lift of the immediate-consequence operator.  It uses predicate-definition embeddings and latent composition channels to form soft body-predicate mixtures, aggregate over all possible witnesses, propose query-conditioned head facts, and update the interpretation through a monotone probabilistic OR.  We instantiate the framework on Knowledge-aware Visual Question Answering (KVQA), and  demonstrates how \ours{} supports end-to-end perceptual grounding, KG evidence injection, and differentiable deductive closure in one trainable architecture.
\end{abstract}

\input{SECTIONS/01_Introduction}
\input{SECTIONS/02_Related_Work}
\input{SECTIONS/03_Technical_Approach}
\input{SECTIONS/04_Experimental_Evaluation}
\input{SECTIONS/05_Conclusions}

\bibliography{refs}

\clearpage\newpage
\pagenumbering{arabic}
\appendix
\input{SECTIONS/appendix}

\end{document}

%% file: macros.tex
\newcommand{\ours}{\textsc{SoftReason}\xspace}

\newcommand{\scallop}{\textsc{Scallop}\xspace}

\newcommand{\kvqa}{KVQA\xspace}

\newcommand{\wikidata}{Wikidata\xspace}

\newcommand{\best}[1]{\ifmmode\boldsymbol{#1}\else\textbf{#1}\fi}

\newcommand{\para}[1]{\par\noindent\textbf{#1}\ }
\newcommand{\eg}{\emph{e.g.},\xspace}
\newcommand{\ie}{\emph{i.e.},\xspace}


%% file: SECTIONS/01_Introduction.tex

\section{Introduction}
\label{sec:intro}

Deep neural networks, especially large language models (LLMs), have achieved
remarkable success in perception and generation tasks (\eg image classification~\citep{dosovitskiy2021vit}, text generation~\citep{openai2023gpt4}).  Given sufficient training data, deep models learn the underlying statistical distribution of the input data and the correlations between input data and target variables. However, learning statistical correlations is not sufficient to perform compositional generalization and deductive reasoning tasks, where logical rules should be applied to known and/or perceived (from input data) facts to derive new conclusions that have not been encountered in the training data or cannot be made from perceptual data ~\citep{lake2018generalization,garcez2019neuralsymbolic,evans2018dilp}.

Neuro-symbolic architectures (\eg DeepProbLog~\citep{manhaeve2018deepproblog},
Scallop~\citep{huang2021scallop}, and Neural Theorem Provers~\citep{rocktaschel2017provers}) address these challenges by \emph{pipelining} neural perception with symbolic reasoning. These systems leverage discrete symbols (predicted by neural networks) and apply logical constraints, rules and search algorithms to perform reasoning tasks, such as multi-hop question answering~\citep{rocktaschel2017provers,saxena2020embedkgqa}, knowledge graph completion~\citep{bordes2013transe,sun2019rotate}, and theorem proving~\citep{rocktaschel2017provers,olausson2023linc}, that cannot be solved using pure neural inductive reasoning.  The fundamental limitation, however, of existing neuro-symbolic pipelines is the gradient gap between neural perception and symbolic reasoning~\citep{manhaeve2018deepproblog,xu2018semanticloss}.  Classical pipelines convert perceptual predictions into discrete symbols (\eg labels and relations) before reasoning begins, introducing non-differentiable boundaries, which does not (1) allow reasoning to shape the latent space learned by perception, (2) allow reasoning to leverage the rich uncertainty (\ie distribution over candidate entities and relations) encoded in learned representations.

We introduce \ours{}, a fully differentiable neuro-\emph{soft}-symbolic deductive reasoning architecture that eliminates the gradient gap between neural perception and symbolic reasoning. A perceptual encoder maps the
high-dimensional perceptual input to dense tokens, a grounding module distributes those tokens over candidate entities, and a relational attention encoder contextualizes the result. \ours{} represents the deductive state as a soft interpretation tensor, which allows the reasoning and perceptions \emph{components} to propagate gradients backward on the computation graph, without an irreversible, hard commitment to discrete symbols at any point in the architecture. The deductive component of the architecture can be trained with any source of deductive rules, including Prolog-like rules or differentiable rule templates. Therefore, deductive reasoning is not a post-processing step but an intrinsic part of what the model learns. Our contributions  are:
\begin{itemize}[leftmargin=1.2em,itemsep=1pt,topsep=2pt]
  \item A formulation of deductive reasoning over perceptual data as
        differentiable closure over a local soft interpretation tensor, which
        preserves uncertainty through every deductive step.
  \item A learned differentiable lift of the immediate-consequence operator
        , implemented through predicate-definition embeddings and latent
        composition channels, with classical Horn-chain reasoning recovered as a
        limiting case.
  \item A fully differentiable end-to-end architecture that unifies
        perceptual grounding, KG evidence injection, and deductive closure in
        one trainable model.
  \item An instantiation of \ours{} on knowledge-aware visual question
        answering.
\end{itemize}

%% file: SECTIONS/02_Related_Work.tex

\section{Related Work}
\label{sec:related}

\para{Neuro-symbolic learning and differentiable logic.}
Neuro-symbolic methods combine neural representation learning with
symbolic knowledge representation and reasoning
(\eg \cite{garcez2019neuralsymbolic}).  Existing systems differ in where they place
the symbolic interface.  Semantic loss converts logical constraints over
structured outputs into a differentiable training objective
(\eg \cite{xu2018semanticloss}).  Logic Tensor Networks interpret first-order logic
(FOL) formulas with many-valued differentiable semantics and use logical
satisfaction as a learning signal~\citep{badreddine2022ltn}.  TensorLog
compiles classes of probabilistic first-order queries into differentiable
functions, making deductive database inference compatible with neural learning
infrastructure~\citep{cohen2016tensorlog}.  These methods show that symbolic
structure can shape neural learning, but the symbolic vocabulary, clauses, or
constraints are usually fixed inputs to the system rather than learned from
perceptual evidence as part of one differentiable deductive state.

\para{Differentiable theorem proving and logic programming.}
Several lines of work make proof search or logic-program inference compatible
with gradient-based learning.  Neural Theorem Provers replace symbolic
unification with differentiable similarity in embedding space and support
multi-hop reasoning over knowledge bases~\citep{rocktaschel2017provers}.
Differentiable Inductive Logic Programming learns soft selections over candidate
rules and can be connected to neural predictors for ambiguous inputs
\citep{evans2018dilp}.  DeepProbLog integrates neural predicates with
probabilistic logic programming, allowing neural predictions to participate in
logical inference~\citep{manhaeve2018deepproblog}.  Scallop builds on
probabilistic deductive databases and provenance semantics to scale
differentiable reasoning over Datalog-style programs~\citep{huang2021scallop}.
These systems substantially narrow the distance between learning and deduction.
However, they still typically rely on a supplied logic program, rule template,
or symbolic proof graph.  The neural component often predicts facts that are
then consumed by a symbolic or compiled reasoning substrate.

\para{Multi-hop reasoning over perceptual data.}
High-dimensional perception make this interface harder because the
premises of a proof are not observed as clean atoms.  Visual reasoning systems
such as Neural-Symbolic Visual Question Answering (VQA) and the
Neuro-Symbolic Concept Learner recover object-centric scene representations and
execute symbolic programs over those representations
\citep{yi2018nsvqa,mao2019nscl}.  This design gives strong compositional
generalization when the scene representation and program trace are correct, but
it also exposes a brittle handoff.  Perception must be converted into a discrete
symbolic state before reasoning, and errors introduced by grounding, parsing,
entity linking, or top-$k$ pruning can remove facts before the deductive module
has a chance to use them.  Similar concerns arise in knowledge-intensive
perceptual reasoning systems that construct explicit graph 
structures before multi-hop inference~\citep{heo2022hypergraph}.

\para{Knowledge-graph completion.}
Knowledge-graph completion methods learn distributed representations of
entities and relations for link prediction over a fixed symbolic graph.
TransE models each relation as a translation in a low-dimensional entity
space, scoring a candidate triple by how closely the head entity shifted by
the relation lands on the tail entity~\citep{bordes2013transe}.  RotatE extends this to
complex vector space, treating each relation as a rotation and thereby
supporting relational path composition and pattern inference~\citep{sun2019rotate}.
ComplEx models asymmetric relations through Hermitian inner products over
complex embeddings~\citep{trouillon2016complex}.  These representations have
been incorporated into multi-hop Knowledge-Graph Question Answering (KGQA)
pipelines, where they score candidate answers after a separate entity-linking
stage~\citep{saxena2020embedkgqa}.  Since these methods operate over a fixed 
graph rather than a differentiable perceptual interpretation, grounding errors
cannot be recovered through answer-level supervision.

\para{LLM-augmented symbolic reasoning.}
A distinct line of work addresses multi-hop reasoning by coupling large language
models (LLMs) with external symbolic solvers.  LINC reformulates logical reasoning as
neurosymbolic programming, where an LLM acts as a semantic parser that
translates natural language premises into first-order logic expressions, which
are then discharged to an external theorem prover~\citep{olausson2023linc}.
Logic-LM follows a similar pipeline, translating questions into symbolic
formulations and using a deterministic solver with an LLM-guided
self-refinement loop~\citep{pan2023logiclm}.  These methods demonstrate strong
performance on formal reasoning benchmarks by leveraging the scale of
pretrained LLMs.  

%% file: SECTIONS/03_Technical_Approach.tex

\section{Technical Approach}
\label{sec:softreason}

We address the problem of deductive reasoning when the input facts (\eg object relations in an image) are latent in high-dimensional perceptual data rather than observed as discrete symbols. Deductive reasoning requires a schema because the system must know which
predicates are available, what argument types they take, and how facts can
compose into new facts.  Knowledge graphs and FOL resources provide this
structure, where  the schemas define the predicate vocabulary, their triples provide
trusted facts, and their relational regularities or explicit clauses indicate
which conclusions may be derived even when the conclusion is not already stored
as a KG triple.  A KG is therefore not treated merely as a lookup table.  It
supplies the symbolic support on which soft deduction is performed. \ours{} combines neural perception to propose probabilistic facts over the local universe, and KG facts to be injected as high-confidence soft evidence rather than as non-differentiable constraints. The resulting state represents facts that are likely observed, facts that are known from the KG, and facts that are deduced from both sources in the same tensor.  KG structure therefore enriches perceptual facts, while the final task loss still propagates back through the perceptual grounding and fact extraction modules.

\subsection{Problem Formulation}
\label{sec:softreason_formulation}

Let $x \in \mathcal{X}$ be a perceptual observation, $q$
denote a query, and let
$\mathcal{K}=(\mathcal{C},\mathcal{P},\mathcal{T})$ be a knowledge source with
discrete symbols $\mathcal{C}$, predicate schema $\mathcal{P}$, and known ground facts
$\mathcal{T}$.  The predicate schema may come from a knowledge graph (KG), or first-order logic (FOL)
definitions.  Although we present the model for binary predicates
$p(c_i,c_j)$, because binary relations cover the common graph setting, unary
predicates can be represented as relations to a distinguished truth constant,
and higher-arity predicates can be represented either with arity-specific
tensors or with standard reification into relation nodes.

\begin{figure}[t]
  \centering
  \includegraphics[width=\textwidth]{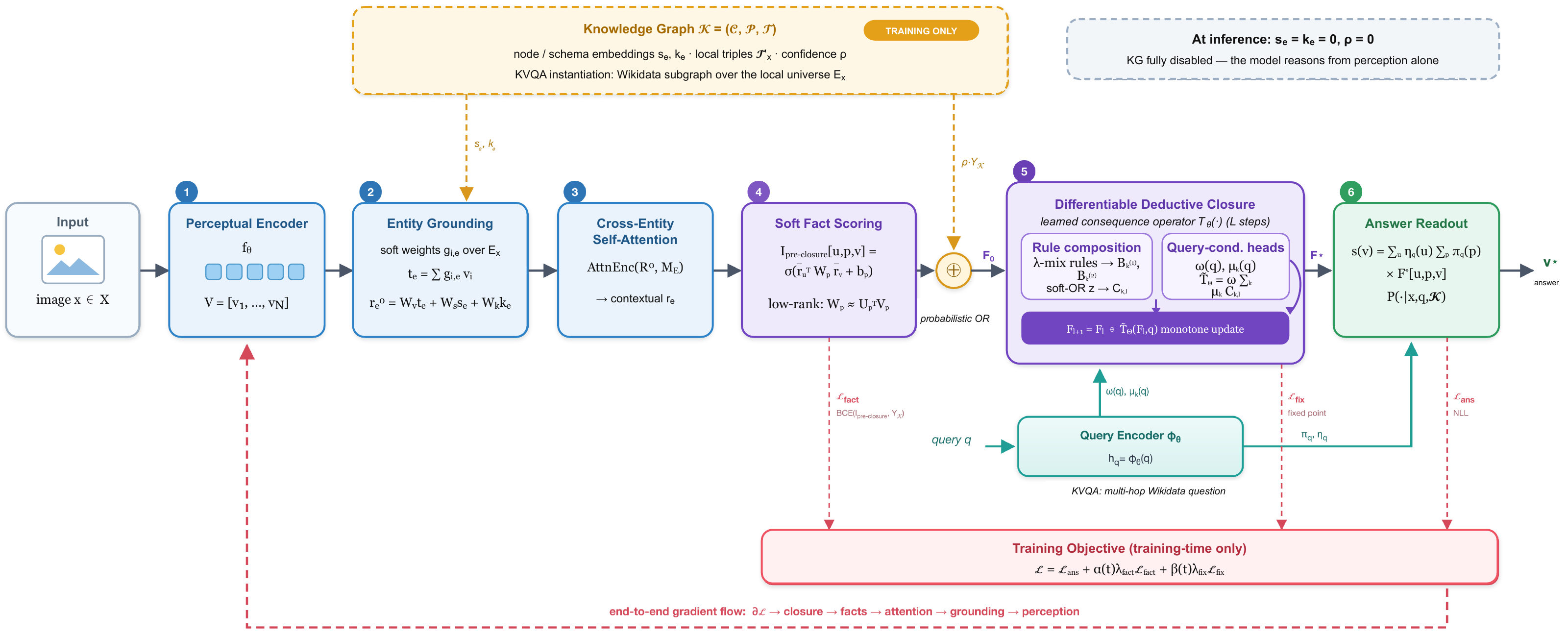}
  \caption{\ours{} architecture, illustrating fully differentiable perception-to-reasoning.}
  \label{fig:softreason_architecture_v3}
\end{figure}

The objective is to map $(x, q, \mathcal{K})$ to an answer distribution through a sequence of fully differentiable stages. All intermediate representations \emph{must} remain soft, and only the final answer selection is mapped to discrete symbols. The fundamental principal of the architecture is that KG triples serve as \emph{training supervision only}. Therefore, during training, KG-node embeddings and KG evidence injection anchor perceptual grounding and fact construction. At inference, on the other hand, both are disabled and the model reasons from perception alone. In the KVQA instantiation, for example, $x$ is an image of a named person, $q$ asks for an answer that can only be obtained by following one or more \wikidata{} relations, and $\mathcal{K}$ is the local \wikidata{} subgraph, as illustrated in \cref{fig:softreason_architecture_v3}.

\subsection{\ours{} Architecture}
\label{sec:softreason_architecture}

For each input $x$, we construct a finite local universe $E_x=\{e_1,\ldots,e_m\}\subseteq\mathcal{C}$ of candidate discrete symbols and a local predicate set $P_x\subseteq\mathcal{P}$. Each predicate $p \in P_x$ is associated with a schema description (or natural-language label) $d(p)$ and is made available to every stage through a learned embedding $a_p$, as shown in \cref{eq:softreason_predicate_definition_embedding},
\begin{equation}
  a_p = \psi_{\mathrm{pred}}\!\left(d(p)\right)
  \in \mathbb{R}^{d_{\mathrm{pre}}},
  \qquad p\in P_x,
  \label{eq:softreason_predicate_definition_embedding}
\end{equation}
where $\psi_{\mathrm{pred}}$ is a learned predicate-definition encoder and $d_{\mathrm{pre}}$ is the predicate-embedding dimension. These embeddings are the \emph{only} schema-specific interface, and the architecture does not assume a particular perception modality, KG source, or rule language, provided input is represented as perceptual tokens, local  symbols, and predicate-definition embeddings.

\para{Perceptual tokenization and entity grounding.}
A perceptual encoder $f_\theta$ maps $x$ to $N$ dense tokens, and the token matrix $V$ is defined as shown in \cref{eq:softreason_visual_tokens},
\begin{equation}
  V = f_\theta(x) = [v_1,\ldots,v_N],
  \qquad v_i \in \mathbb{R}^{d_{\mathrm{per}}}.
  \label{eq:softreason_visual_tokens}
\end{equation}
where $d_{\mathrm{per}}$ is the perceptual encoder dimension. A grounding head distributes each token over the  discrete symbols $E_x$, and the grounding weight $g_{i,e}$ is computed as shown in \cref{eq:softreason_grounding},
\begin{equation}
  g_{i,e}
  =
  \frac{\exp(\ell_{i,e})}
       {\sum_{e'\in E_x}\exp(\ell_{i,e'})},
  \qquad
  \ell_{i,\cdot}=h_\theta(v_i),
  \label{eq:softreason_grounding}
\end{equation}
where $h_\theta$ is a learned projection, $i \in \{1,\ldots,N\}$ indexes perceptual tokens, and $e \in E_x$ indexes local discrete symbols. Each local entity token $t_e$ is the grounding-weighted perceptual summary, as shown in \cref{eq:softreason_entity_token},
\begin{equation}
  t_e = \sum_{i=1}^{N} g_{i,e}\, v_i.
  \label{eq:softreason_entity_token}
\end{equation}
The initial entity representation $r_e^0$ fuses the perceptual token with KG-node and schema embeddings, and $r_e^0$ is defined as shown in \cref{eq:softreason_entity_init},
\begin{equation}
  r_e^0 = W_v t_e + W_s s_e + W_k k_e
  \in \mathbb{R}^{d_{\mathrm{ent}}},
  \label{eq:softreason_entity_init}
\end{equation}
where $W_v$, $W_s$, and $W_k$ are learned projection matrices, $d_{\mathrm{ent}}$ is the entity-representation dimension, and $s_e$ and $k_e$ are looked up from the knowledge source $\mathcal{K}$ for entity $e$: $s_e$ is a schema or textual embedding and $k_e$ is a KG-node embedding. Both terms act as a training scaffold, providing a KG-grounded initialization that helps the model associate entity representations with known relational structure. At inference, both are set to zero, reducing \cref{eq:softreason_entity_init} to $r_e^0 = W_v t_e$.

\para{Cross-entity contextualization.}
Multi-head self-attention contextualizes the entity representations  by aggregating evidence across all candidates in $E_x$. Given initial tokens $R^0 = \{r^0_e : e \in E_x\}$, the contextualized matrix $R$ is computed as shown in \cref{eq:softreason_attention},
\begin{equation}
  R = \operatorname{AttnEnc}_\theta(R^0, M_E),
  \label{eq:softreason_attention}
\end{equation}
where  $M_E$ is a binary mask that excludes padded entities. The resulting $r_e$ captures pairwise relational context needed to score candidate facts.

\para{Soft fact scoring and KG injection.}
Using normalized representations $\bar{r}_e = \operatorname{LayerNorm}(r_e)$, a predicate-conditioned bilinear head scores each candidate triple, and the pre-closure score $I_{\text{pre-closure}}[u,p,v]$ is defined as shown in \cref{eq:softreason_bilinear_fact},
\begin{equation}
  I_{\text{pre-closure}}[u,p,v]
  =
  \sigma\!\left(\bar r_u^\top W_p \bar r_v + b_p\right),
  \label{eq:softreason_bilinear_fact}
\end{equation}
where $u,v$ index local discrete symbols in $E_x$, $p$ indexes predicates in $P_x$, $\bar r_u, \bar r_v \in \mathbb{R}^{d_{\mathrm{ent}}}$, $W_p \in \mathbb{R}^{d_{\mathrm{ent}} \times d_{\mathrm{ent}}}$ is a predicate-specific weight matrix, $b_p$ is a scalar bias, and $\sigma$ is the logistic sigmoid. The predicate matrix admits a low-rank factorization, as shown in \cref{eq:softreason_low_rank_fact},
\begin{equation}
  W_p \approx U_p^\top V_p,
  \qquad U_p, V_p \in \mathbb{R}^{r \times d_{\mathrm{ent}}}.
  \label{eq:softreason_low_rank_fact}
\end{equation}
where $r \ll d_{\mathrm{ent}}$ is the rank, reducing parameters from $O(|P_x|d_{\mathrm{ent}}^2)$ to $O(2|P_x|r\,d_{\mathrm{ent}})$. During training, known KG triples form a binary target tensor, as shown in \cref{eq:softreason_kg_target},
\begin{equation}
  Y_{\mathcal{K}}[u,p,v]
  =
  \mathbf{1}\!\left[(e_u, p, e_v) \in \mathcal{T}_x\right],
  \label{eq:softreason_kg_target}
\end{equation}
where $\mathbf{1}[\cdot]$ is the indicator function and $\mathcal{T}_x \subseteq \mathcal{T}$ denotes KG triples restricted to $E_x$. The initial interpretation $F_0$ merges neural scores with KG evidence via a probabilistic OR, and $F_0$ is defined as shown in \cref{eq:softreason_evidence_injection},
\begin{equation}
  F_0
  =
  I_{\text{pre-closure}} \oplus \rho Y_{\mathcal{K}}
  =
  1 - (1-I_{\text{pre-closure}})(1-\rho Y_{\mathcal{K}}),
  \qquad 0 < \rho < 1,
  \label{eq:softreason_evidence_injection}
\end{equation}
where $\rho \in (0,1)$ is a scalar KG confidence weight and $\oplus$ denotes the probabilistic OR. Known KG facts enter the reasoning state with high confidence while gradients still propagate through $I_{\text{pre-closure}}$. At inference, $\rho = 0$, therefore $F_0 = I_{\text{pre-closure}}$.

\para{Soft rule composition.}
The deductive state is a soft \emph{semantic interpretation tensor}
\begin{equation}
  F \in [0,1]^{m \times |P_x| \times m},
  \qquad
  F[u,p,v] \approx \Pr\!\left[p(e_u,e_v)\mid x,\mathcal{K}\right],
  \label{eq:softreason_interpretation}
\end{equation}
where $m=|E_x|$, $u, v \in \{1,\ldots,m\}$ index entities in $E_x$, and $p \in \{1,\ldots,|P_x|\}$ indexes predicates in $P_x$. The classical Boolean consequence operator $T_P$ applies $\bigwedge$ over body atoms and $\bigvee$ over matching rule instantiations. Replacing these with a differentiable t-norm $\otimes$ and a soft-or yields, for a binary Horn rule $p_1(x,z) \wedge p_2(z,y) \rightarrow p_h(x,y)$, the relaxed consequence score $\widetilde{T}_P(F)[x,p_h,y]$ shown in \cref{eq:softreason_prob_rule},
\begin{equation}
  \widetilde{T}_P(F)[x,p_h,y]
  =
  \operatorname*{soft\text{-}or}_{z}
  \!\left(F[x,p_1,z] \otimes F[z,p_2,y]\right),
  \label{eq:softreason_prob_rule}
\end{equation}
where $\otimes$ is a differentiable t-norm and the soft-or marginalizes over witness (\ie existentially quantified) constants $z$.

We generalize \cref{eq:softreason_prob_rule} by learning body predicates and witness aggregation through $K$ latent composition channels, such that, for channel $k$, trainable body keys $\alpha_k, \beta_k \in \mathbb{R}^{d_{\mathrm{pre}}}$ define soft predicate mixtures over $P_x$, and the mixture weights $\lambda^{(1)}_{p,k}$ and $\lambda^{(2)}_{p,k}$ are defined as:
\begin{equation}
  \lambda^{(1)}_{p,k}
  =
  \frac{\exp(a_p^\top \alpha_k)}
       {\sum_{p'\in P_x}\exp(a_{p'}^\top \alpha_k)},
  \qquad
  \lambda^{(2)}_{p,k}
  =
  \frac{\exp(a_p^\top \beta_k)}
       {\sum_{p'\in P_x}\exp(a_{p'}^\top \beta_k)}.
  \label{eq:softreason_body_mixtures}
\end{equation}
The two soft body relations for channel $k$ at step $\ell$ are calculated as shown in \cref{eq:softreason_soft_bodies},
\begin{equation}
  B^{(1)}_{k,\ell}[u,z]
  =
  \sum_{p\in P_x}\lambda^{(1)}_{p,k} F_\ell[u,p,z],
  \qquad
  B^{(2)}_{k,\ell}[z,v]
  =
  \sum_{p\in P_x}\lambda^{(2)}_{p,k} F_\ell[z,p,v],
  \label{eq:softreason_soft_bodies}
\end{equation}
where $F_\ell$ is the interpretation tensor at reasoning step $\ell$. The witness-marginalized composition score is shown in \cref{eq:softreason_noisy_or_witness},
\begin{equation}
  C_{k,\ell}[u,v]
  =
  1
  -
  \exp\!\left(
    \frac{1}{m}
    \sum_{z\in E_x}
    \log\!\left(
      1 - \operatorname{clip}_{\epsilon}
      \!\left(B^{(1)}_{k,\ell}[u,z] \cdot B^{(2)}_{k,\ell}[z,v]\right)
    \right)
  \right),
  \label{eq:softreason_noisy_or_witness}
\end{equation}
where $\epsilon>0$ is a small constant and $\operatorname{clip}_\epsilon(\cdot)$ clamps to $[\epsilon, 1{-}\epsilon]$ for numerical stability.

\para{Query-conditioned closure.}
The query embedding $h_q = \phi_\theta(q)$ selects which head predicates are relevant and how channels are weighted. For each candidate head predicate $p_h$, the query-conditioned channel mixture $\mu_{k,p_h}(q)$ is shown in \cref{eq:softreason_head_mix},
\begin{equation}
  \mu_{k,p_h}(q)
  =
  \frac{\exp(a_{p_h}^{\top}\gamma_k + (W_r h_q)_k + c_k)}
       {\sum_{k'=1}^{K}\exp(a_{p_h}^{\top}\gamma_{k'} + (W_r h_q)_{k'} + c_{k'})},
  \label{eq:softreason_head_mix}
\end{equation}
where $\gamma_k \in \mathbb{R}^{d_{\mathrm{pre}}}$ is a trainable head-key vector for channel $k$, $W_r$ is a learned query projection, and $c_k$ is a learned scalar bias. A head gate selectively activates each candidate head predicate, and the gate value $\omega_{p_h}(q)$ is shown in \cref{eq:softreason_head_gate},
\begin{equation}
  \omega_{p_h}(q)
  =
  \sigma\!\left((W_h h_q)^\top a_{p_h}\right).
  \label{eq:softreason_head_gate}
\end{equation}
where $W_h$ is a learned projection matrix. The one-step learned consequence proposal is: 
\begin{equation}
  \widetilde{T}_{\Theta}(F_\ell,q)[u,p_h,v]
  =
  \omega_{p_h}(q)
  \sum_{k=1}^{K}
  \mu_{k,p_h}(q)\, C_{k,\ell}[u,v],
  \label{eq:softreason_learned_tp}
\end{equation}
where $\Theta$ denotes all learned parameters. When $\lambda^{(1)}$, $\lambda^{(2)}$, and $\mu$ collapse to one-hot selections, \cref{eq:softreason_learned_tp} recovers a selected Horn-chain rule, while during training it remains a smooth operator over the full predicate schema. Facts accumulate across reasoning steps through a monotone probabilistic OR update, as shown in \cref{eq:softreason_monotone_update},
\begin{equation}
  F_{\ell+1}
  =
  F_\ell \oplus \widetilde{T}_{\Theta}(F_\ell,q)
  =
  1 - (1-F_\ell)(1-\widetilde{T}_{\Theta}(F_\ell,q)),
  \label{eq:softreason_monotone_update}
\end{equation}
This update mirrors least-fixed-point forward chaining, since $F_{\ell+1} \geq F_\ell$ cell-wise ensures the sequence is bounded and non-decreasing. After $L$ iterations the final closed interpretation is: 
\begin{equation}
  F^\star = F_L
  =
  \left(\widetilde{T}_{\Theta}^{\oplus}\right)^L\!(F_0,q),
  \label{eq:softreason_closure}
\end{equation}
where $\widetilde{T}_{\Theta}^{\oplus}$ denotes one consequence proposal (\cref{eq:softreason_learned_tp}) composed with the probabilistic OR update (\cref{eq:softreason_monotone_update}), and $L$ controls maximum deductive depth. Since all operations are differentiable, the answer loss propagates gradients back through closure, fact extraction, grounding, and perception in a single pass.

\para{Answer readout.}
The query predicate is not required to be a hard label. \ours{} computes a soft query predicate distribution, and $\pi_q(p)$ is defined as shown in \cref{eq:softreason_query_predicate},
\begin{equation}
  \pi_q(p) = \operatorname{softmax}(W_q h_q)_p,
  \label{eq:softreason_query_predicate}
\end{equation}
where $W_q$ is a learned projection matrix. A query anchor distribution $\eta_q(u)$ is defined over local discrete symbols, and it is one-hot when the anchor is known. The unnormalized answer score $s(v)$ for candidate $v$ is shown in \cref{eq:softreason_answer_score},
\begin{equation}
  s(v)
  =
  \sum_{u\in E_x} \eta_q(u)
  \sum_{p\in P_x} \pi_q(p)\, F^\star[u,p,v],
  \label{eq:softreason_answer_score}
\end{equation}
and the training answer distribution $P_{\Theta}(v\mid x,q,\mathcal{K})$ is the normalized masked score:
\begin{equation}
  P_{\Theta}(v\mid x,q,\mathcal{K})
  =
  \frac{M_E(v)\left(s(v)+\epsilon\right)}
       {\sum_{v'\in E_x}M_E(v')\left(s(v')+\epsilon\right)},
  \label{eq:softreason_answer_distribution}
\end{equation}
where $M_E(v) \in \{0,1\}$ masks padded discrete symbols and $\epsilon > 0$ prevents zero-probability targets during early training. For a query anchored at $e_a$ with predicate $p_q$, the answer set read from $F^\star$ is shown in \cref{eq:softreason_answer_set},
\begin{equation}
  \mathcal{A}(x,q)
  =
  \{e_b \in E_x : F^\star[a,p_q,b] \text{ is high}\}.
  \label{eq:softreason_answer_set}
\end{equation}
where $a$ and $b$ are indices such that $e_a, e_b \in E_x$.

\para{Training objective.}
The  architecture is trained with three jointly optimized losses. The answer loss supervises the final readout against the true answer $y \in E_x$, as shown in \cref{eq:softreason_answer_loss},
\begin{equation}
  \mathcal{L}_{\mathrm{ans}}
  =
  -\log P_{\Theta}(y\mid x,q,\mathcal{K}).
  \label{eq:softreason_answer_loss}
\end{equation}
The fact loss supervises the pre-closure extractor against known KG facts with class balancing, as shown in \cref{eq:softreason_fact_loss},
\begin{equation}
  \mathcal{L}_{\mathrm{fact}}
  =
  \frac{
    \sum_{u,p,v}
    M_E(u)M_E(v)\, w_{u,p,v}\,
    \operatorname{BCE}\!\left(I_{\text{pre-closure}}[u,p,v],Y_{\mathcal{K}}[u,p,v]\right)
  }{
    \sum_{u,p,v} M_E(u)M_E(v)\, w_{u,p,v}
  },
  \label{eq:softreason_fact_loss}
\end{equation}
where $u,v$ index local discrete symbols, $p$ indexes predicates, and $w_{u,p,v} > 0$ assigns larger weights to positive KG triples to compensate for label sparsity. The binary cross-entropy is defined as $\operatorname{BCE}(\hat{y},y) = -y\log\hat{y} - (1-y)\log(1-\hat{y})$. The fixed-point loss penalizes residual change after one additional closure step beyond $F^\star$, as shown in \cref{eq:softreason_fix_loss},
\begin{equation}
  \mathcal{L}_{\mathrm{fix}}
  =
  \frac{
    \sum_{u,p,v} M_E(u)M_E(v)
    \!\left(
      F^\star[u,p,v]
      -
      \bigl(F^\star \oplus \widetilde{T}_{\Theta}(F^\star,q)\bigr)[u,p,v]
    \right)^{\!2}
  }{
    \sum_{u,p,v} M_E(u)M_E(v)
  }.
  \label{eq:softreason_fix_loss}
\end{equation}
The complete objective is $
  \mathcal{L}
  =
  \mathcal{L}_{\mathrm{ans}}
  +
  \alpha(t)\lambda_{\mathrm{fact}}\mathcal{L}_{\mathrm{fact}}
  +
  \beta(t)\lambda_{\mathrm{fix}}\mathcal{L}_{\mathrm{fix}},
  \label{eq:softreason_total_loss}
$
where $t$ is the training step, $\lambda_{\mathrm{fact}}, \lambda_{\mathrm{fix}} > 0$ are fixed weights, and $\alpha(t)$, $\beta(t)$ are optional curriculum schedules over $t$. The answer loss teaches the model which closure facts solve the query, the fact loss anchors the perceptual extractor to  KG evidence, and the fixed-point loss encourages the learned consequence operator to converge as a soft deductive closure.

\paragraph{Inference} At inference, the KG $\mathcal{K}$ is not used and KG-node and schema embeddings are disabled ($s_e = k_e = \mathbf{0}$), so \cref{eq:softreason_entity_init} reduces to $r_e^0 = W_v t_e$.  The KG evidence injection is also disabled ($\rho = 0$), so the initial interpretation simplifies to $F_0 = I_{\text{pre-closure}}$.  All remaining stages, operate identically to training.  The predicted answer is $v^\star = \arg\max_{v \in E_x} s(v)$.

%% file: SECTIONS/04_Experimental_Evaluation.tex

\section{Experimental Evaluation}
\label{sec:experiments}

\paragraph{Benchmark.}
\input{SECTIONS/TBL_01.tex}
We use Knowledge-aware Visual Question Answering (KVQA) for evaluating the performance of \ours{} on multi-hop reasoning over perceptual data. It is important to note that the goal is not to evaluate
\ours{} as a generic Visual Question Answering (VQA) model.  The goal is to
test whether a fully differentiable neuro-soft-symbolic architecture can ground
perceptual evidence, inject Knowledge Graph (KG) evidence, and learn deductive
closure over KG-provided predicates.  KVQA~\citep{shah2019kvqa} is a useful
testbed because each example starts from a named person in an image and asks for
an answer obtained by following one or more \wikidata{} relations.

\para{Evaluation setting.}
We adopt the \emph{entity-linking} setting, where the model has to
ground the image to a distribution over named entities before reasoning.  This
setting is the direct test of our claim because \ours{} keeps the path from
perception to deductive closure differentiable.  Gradients from the answer loss
flow through the learned closure distribution, KG fact supervision, attention
layers, soft fact construction, and perceptual grounding heads.

\para{Metrics and baselines.}
We use Hit@1, the closed-vocabulary answer accuracy used by
\kvqa{}, and Recall@5 to align with ranking-based
 evaluations such as \scallop{}~\citep{huang2021scallop}.
We use \kvqa{} baselines from ~\citet{shah2019kvqa} and
the Hypergraph Transformer study~\citep{heo2022hypergraph}.  We also include ~\citet{garciaolano2022entity}, whose entity-enhanced knowledge injection
method uses automatic named entity recognition on the question text and reports
results on ~\citet{shah2019kvqa}'s official five splits.

\para{KVQA Entity-Linking Evaluation}
\label{sec:exp_main_kvqa}
\cref{tab:kvqa_entity_linking} compares
\ours{} against published \kvqa{} models under the realistic entity-linking
setting. The significant improvements over prior methods suggest that
deductive supervision shapes visual grounding and fact construction when the
reasoning operator remains differentiable.

\para{Reasoning Evaluation.}
\label{sec:exp_reasoning_slices}
Aggregate answer accuracy is not sufficient to validate the deductive reasoning capabilities.
A one-hop question can often be answered by entity grounding and direct fact retrieval,
while two- and three-hop questions require relational composition over KG
predicates.  We therefore report hop-depth results for 1-hop, 2-hop, 3-hop, and
the combined multi-hop subset. \cref{tab:kvqa_hop_breakdown} shows the breakdown of \ours{}'s performance by hop depth.
Improvements on higher-hop subsets instead provide evidence that the learned differentiable closure operator contributes to deductive composition.
\input{SECTIONS/TBL_03.tex}

\para{Why no LLM-augmented or OK-VQA baselines?} LLM-augmented methods (\eg LINC~\citep{olausson2023linc} and Logic-LM~\citep{pan2023logiclm})  do not learn a differentiable perceptual grounding mechanism. Further, to the best of our knowledge, there are no published results on KVQA for these methods. The knowledge in OK-VQA~\citep{okvqa} is unstructured text and the task is to connect images to relevant text.  \ours{} is designed to learn differentiable deductive closure over a structured knowledge graph.

%% file: SECTIONS/TBL_01.tex
\begin{wraptable}{r}{0.75\linewidth}
\centering
\small
\caption{\ours{} performance on KVQA. Baselines use an external entity linker, while \ours{} is end-to-end perception to reasoning. Results are from the full test split.  Hit@1 indicates closed-vocabulary accuracy. R@5 indicates ranked recall, reported to align with Scallop-style evaluations~\citep{huang2021scallop}. $^{\star}$Baselines reported in \citet{shah2019kvqa}. $^{\dagger}$MemNN with additional supervision. ~\citet{garciaolano2022entity}~report the best fully automatic NER variant (NERagro, noisy) averaged over \citet{shah2019kvqa}'s five official splits. N/A indicates the original paper did not report this metric.}
\label{tab:kvqa_entity_linking}
\begin{tabular}{l c c}
\toprule
\textbf{Method} & \textbf{Hit@1} & \textbf{R@5} \\
\midrule
\rowcolor{gray!15}
\multicolumn{3}{l}{\textbf{External entity linker}} \\
\hspace{0.75em}BLSTM$^{\star}$ & 37.60 & N/A \\
\hspace{0.75em}MemNN$^{\star}$ & 42.20 & N/A \\
\hspace{0.75em}GCN$^{\star}$ & 48.50 & N/A \\
\hspace{0.75em}GGNN$^{\star}$ & 50.90 & N/A \\
\hspace{0.75em}MemNN$^{\star\dagger}$ & 54.00 & N/A \\
\hspace{0.75em}HAN$^{\star}$ & 53.30 & N/A \\
\hspace{0.75em}BAN$^{\star}$ & 59.80 & N/A \\
\hspace{0.75em}Transformer (SA)$^{\star}$ & 58.30 & N/A \\
\hspace{0.75em}Transformer (SA+GA)$^{\star}$ & 60.10 & N/A \\
\hspace{0.75em}Hypergraph Transf.~\citep{heo2022hypergraph} & 62.40 & N/A \\
\hspace{0.75em}Garcia-Olano et al.~\citep{garciaolano2022entity} & 50.77 & N/A \\
\midrule
\rowcolor{gray!20}
\multicolumn{3}{l}{\textbf{End-to-end (ours)}} \\
\rowcolor{gray!10}\hspace{0.75em}\textbf{\ours{}} & \textbf{94.30} & \textbf{99.38} \\
\bottomrule
\end{tabular}
\vspace{-.15in}
\end{wraptable}

%% file: SECTIONS/TBL_03.tex
\begin{table}[hb]
\centering
\small
\caption{SoftReason hop-depth metrics. The test split contains no 3-hop examples.}
\label{tab:kvqa_hop_breakdown}
\begin{tabular}{l c c c c c c}
\toprule
\textbf{Method} & \textbf{Hit@1} & \textbf{R@5} & \textbf{1-hop} & \textbf{2-hop} & \textbf{3-hop} & \textbf{multi-hop} \\
\midrule
\ours{ (full test)} & 94.30 & 99.38 & 93.24 & 98.20 & N/A & 98.20 \\
\bottomrule
\end{tabular}
\vspace{2pt}
\end{table}

%% file: SECTIONS/05_Conclusions.tex

\section{Conclusion}
\label{sec:conclusions}

We presented \ours{}, a fully differentiable neuro-soft-symbolic architecture
for deductive reasoning over high-dimensional perceptual data.   Instead of producing discrete symbols before reasoning begins, we represent the deductive state as a soft semantic interpretation tensor, whose cells are differentiable representations of ground symbols.  The core of the architecture is a learned
differentiable lift of the immediate-consequence operator.  It uses predicate-definition embeddings and latent composition channels to perform soft body-predicate mixture, aggregation over candidate witnesses, and monotone
closure update through a probabilistic OR. Horn-chain reasoning is recovered as a limiting case. During training, every predicate choice, witness entity, and derived fact remains connected to the answer loss through the gradient path, which (1) allows deductive reasoning to shape perceptual representations, (2) allows reasoning to leverage the rich uncertainty encoded by learned perceptual representations. We instantiated \ours{} on KVQA benchmark and demonstrated strong improvements over prior knowledge-based VQA methods on the entity-linking protocol, where no oracle entity is supplied.  

%% file: SECTIONS/appendix.tex

\section*{Supplementary Material}
\label{sec:appendix}

\subsection*{Implementation Details}
\label{sec:appendix_impl_details}

For the KVQA instantiation we use an entity budget of $m=20$ candidate
constants per sample, a local predicate set of $|P_x|=100$ \wikidata{}
relations, $K=8$ latent composition channels, and a closure depth of $L=3$
iterations.  The perceptual encoder is a frozen Vision Transformer (ViT-B/16)
\citep{dosovitskiy2021vit}.  The relational attention encoder uses two
Transformer layers with hidden dimension $d=256$.  We train with the Adam
optimizer at a learning rate of $10^{-4}$, a batch size of 32, and loss
weights $\lambda_{\mathrm{fact}}=1.0$ and $\lambda_{\mathrm{fix}}=0.5$ without
curriculum scheduling ($\alpha(t)=\beta(t)=1$).  All experiments run on a
single NVIDIA A100 Graphics Processing Unit (GPU) for 20 epochs.

\subsection*{Complexity and locality}
\label{sec:complexity_locality}
\ours{} reasons over a local universe rather than the full global KG because
the dense interpretation has size $O(m^2|P_x|)$ per sample.  The bilinear fact
head is computed in predicate chunks, and the witness aggregation in
\cref{eq:softreason_noisy_or_witness} is computed in witness-entity chunks.
For $K$ latent composition channels, one learned consequence step has dominant
costs $O(m^2|P_x|K)$ for predicate mixtures and $O(Km^3)$ for soft witness
composition.  These costs are the differentiable analogue of enumerating
possible rule bodies and witness substitutions, but they preserve gradients
through all alternatives instead of materializing a single symbolic proof path.


\subsection*{Other Comparisons}
\label{sec:appendix_other_comparisons}

The methods in this section address Knowledge-aware Visual Question Answering
(KVQA) on the Shah et al. 2019 dataset but are not directly comparable to the
primary entity-linking results in \cref{tab:kvqa_entity_linking}.  We include
them for completeness so that reviewers have a full picture of published results
on this benchmark.

\para{Entity-linking, dataset-label-guided, and oracle protocols.}
It is important to distinguish three experimental protocols used across \kvqa{}
papers.  In the \emph{entity-linking} protocol, the model receives only the raw
image and question and must automatically detect the named entity in the image
through face recognition or visual grounding, link it to a \wikidata{} entry,
and then perform multi-hop reasoning.  This is the protocol used in
\cref{tab:kvqa_entity_linking} and throughout the main evaluation of \ours{}.
In the \emph{dataset-label-guided} protocol, the KVQA annotation files include
the person's name as metadata, and the model uses that name string to retrieve
relevant Knowledge Graph (KG) triples via embedding similarity.  Face
recognition is bypassed, but the model still performs question understanding and
relational reasoning; the name-to-KG matching step replaces the visual grounding
step.  

\para{Dataset-label-guided results.}
Jhalani et al.~\citep{jhalani2024precision} use the dataset-label-guided
protocol.  Their system augments an OFA encoder-decoder with dynamic KG triple
extraction conditioned on the dataset-provided entity name.  Because they use
the person's name from the annotation file rather than automatic face
recognition, their results are not directly comparable to the entity-linking
setting in \cref{tab:kvqa_entity_linking}.  We report their numbers here for
completeness.

\begin{table}[h]
\centering
\small
\caption{KVQA results under the dataset-label-guided protocol~\citep{jhalani2024precision}.
The model uses annotation-provided named entity labels to retrieve KG triples;
face recognition is not required. These results are not directly comparable
to the entity-linking results in \cref{tab:kvqa_entity_linking}.}
\label{tab:kvqa_label_guided}
\begin{tabular}{l l c}
\toprule
\textbf{Method} & \textbf{Venue} & \textbf{Hit@1} \\
\midrule
Jhalani et al.~(Single-Hop) & ICON 2024 & 83.15 \\
Jhalani et al.~(Multi-Hop)  & ICON 2024 & 85.19 \\
\bottomrule
\end{tabular}
\end{table}